\documentclass[twoside,11pt]{article}

\usepackage[nopub]{jmlr2e_modify}

\usepackage{amsmath,graphicx}
\usepackage{mathtools}
\usepackage{algorithm}
\usepackage[]{algpseudocode}
\usepackage[para,flushleft]{threeparttable}
\usepackage{enumitem}
\usepackage{hhline}
\usepackage{multirow}

\jmlrheading{}{2019}{}{1/19}{}{14-115}{Giovanni Di Gennaro, Amedeo Buonanno and Francesco A.N. Palmieri}

\ShortHeadings{Optimized Realization of Bayesian Nets in RN Form using LVM}{Di Gennaro, Buonanno and Palmieri}
\firstpageno{1}

\begin{document}

\title{Optimized Realization of Bayesian Networks \\  in Reduced Normal Form using Latent Variable Model }

\author{\name Giovanni Di Gennaro \email giovanni.digennaro@unicampania.it \\
	   \name Amedeo Buonanno \email amedeo.buonanno@unicampania.it \\
	   \name Francesco A.N.\ Palmieri \email francesco.palmieri@unicampania.it \\
       	   \addr  Universit\`a degli Studi della Campania ``Luigi Vanvitelli"\\
		   Dipartimento di Ingegneria \\
       	   	   via Roma 29, Aversa (CE), Italy}

\editor{}

\maketitle

\begin{abstract}%
	Bayesian networks in their Factor Graph Reduced Normal Form (FGrn) are a powerful paradigm for implementing inference graphs. 	Unfortunately, the computational and memory costs of these networks may be considerable, even for relatively small networks, 	
	and this is one of the main reasons why these structures have often been underused in practice. In this work, through a detailed 
	algorithmic and structural analysis, various solutions for cost reduction are proposed. An online version of the classic batch 
	learning algorithm is also analyzed, showing very similar results (in an unsupervised context); which is essential even if multilevel 
	structures are to be built. The solutions proposed, together with the possible online learning algorithm, are included in a C++ 
	library that is quite efficient, especially if compared to the direct use of the well-known sum-product and Maximum Likelihood (ML) 
	algorithms. The results are discussed with particular reference to a Latent Variable Model (LVM) structure.
\end{abstract}

\begin{keywords}
	Bayesian Networks, Belief Propagation, Factor Graphs, Latent Variable, Optimization
\end{keywords}

\section{Introduction}
The  Factor Graph (FG) representation, and in particular the so-called Normal Form (FGn) \citep{Forney2001, Loeliger2004}, is a very appealing formulation to visualize and manipulate Bayesian graphs; representing their relative joint probability by assigning variables to arcs and functions (or factors) to nodes. Furthermore, in the Factor Graph in Reduced Normal Form (FGrn), through the use of replicator units (or equal constraints), the graph is reduced to an architecture in which each variable is connected to two factors at most \citep{Palmieri2016}; with belief messages that flow bidirectionally into the network. This paradigm has demonstrated its extensive modularity and flexibility in managing variables of different types and cardinalities \citep{PalmBuon2015}, and can also be used to build multi-layer network \citep{Palmieri2014, BuonPalm2015_2, BuonPalm2015_3}.

In a previous work, a Simulink library for the rapid prototyping of an FGrn network was already implemented \citep{BuonPalm2015}, but because of the limitations imposed by Simulink, it is not particularly suitable for treating large amounts of data, and/or complex architectures with many variables. 

Truthfully, despite the large presence of such arguments in the literature \citep{Koller2009, Barber2012, Murphy2012}, this type of structure always suffers from high computational and memory costs, due to the lack of attention given to the specific algorithmic implementation. Therefore, this work aims to improve complexity efficiency in both inference and learning (overcoming software limitations), and all the solutions obtained have been included in a C++ library (\url{https://github.com/mlunicampania/FGrnLib}). This manuscript represents an extended version of the work presented at IEEE International Workshop on Machine Learning for Signal Processing 2018 \citep{MLSP2018}.

The various problems, related to the propagation and learning of probabilities within the FGrn paradigm, are addressed by focusing on the implementation of the Latent Variable Model (LVM) \citep{Bishop1999, Murphy2012}, also called Autoclass \citep{Cheeseman1996}. LVMs can be used in a large number of applications and can be seen as a basic building block for more complex architectures.

After a brief introduction of the FGrn paradigm  in \autoref{sec:rnf} and the LVM  in \autoref{sec:LVM}, necessary to provide the fundamental elements for subsequent discussions, the C++ library project is described  in \autoref{sec:library}. In \autoref{sec:complexity} a detailed analysis of the computational complexity of the various algorithmic elements is presented for  each bulding block. In \autoref{sec:performance} some simulation results that verify  how the proposeed algorithms produce indisputable advantages are presented. Finally, in \autoref{sec:incremental}, an incremental learning algorithm is introduced by modifying the ML recursions, that not only presents a significant decrease in terms of memory costs (allowing learning even in the presence of huge datasets),  but shows in some cases  better performance in avoiding local minima traps.

\begin{figure} [!t]
	\centering
	\vspace{-1em}
	\includegraphics[width=0.6\linewidth]{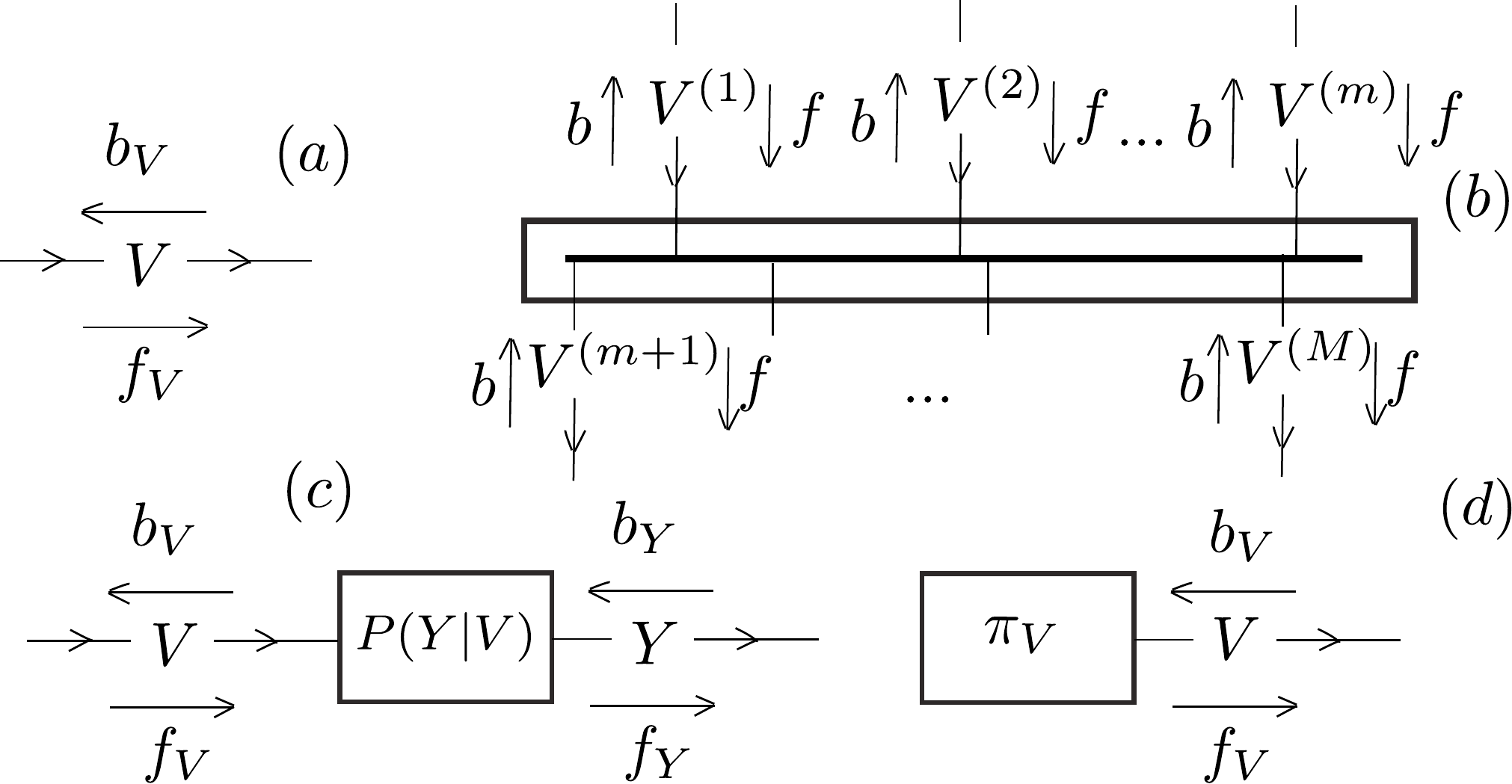}
	\caption{FGrn components: (a) a Variable with the associated forward and backward messages; (b) a Diverter representing the 	
	replication of $M$ variables; (c) a SISO block displaying the conditional distribution matrix of the connected variables; (d) a Source 
	with the prior distribution for the relative variable.}
	\label{fig:blocks}
	\vspace{-1em}
\end{figure}

\section{Factor Graphs in Reduced Normal Form}
\label{sec:rnf}
A FGrn requires only the combination of the elements  shown in \autoref{fig:blocks}: 
\begin{enumerate}[label=(\alph*)]
	\item The {\em Variable} $V$, which can take a single discrete value $v\in{\cal V}=\{ v_{1},...,v_{|{\cal V}|}\}$, is 
	represented 	as an oriented edge with two discrete messages, proportional ($\propto$) to distributions, that travel  in both 
	directions. 
	
	Depending on  the direction assigned to the variable, the two messages are respectively called {\em 
	forward} $f_V(v)$ and {\em backward} $b_V(v)$, and can be also represented as $|{\cal V}|$\nobreakdash-dimensional 
	column vectors: ${\bf f}_V$ and ${\bf b}_V$. Note that the {\em Marginal Distribution} $p_V(v)$, which is proportional to the 
	posterior given the observations 	anywhere 
	else in the network, is proportional to the product 
	\begin{equation*} 
		p_V(v) \propto f_V(v) b_V(v),
	\end{equation*}
	or in vector form 
	\begin{equation} 
		{\bf p}_V \propto {\bf f}_V  \odot {\bf b}_V,
		\label{eq:marginal}
	\end{equation}
	where $\odot$ denotes the Hadamard (element-by-element) product.

	\item The {\em Replicator Block} (Diverter) represents the equality constraint of the connected variables. The constraint of 
	equality between the variables is obtained by making sure that each replica can carry different forward and backward messages.
	A replicator acts like a bus with messages combined and diverted towards the connected branches. The combination rule 
	(product rule) is such that outgoing messages are the product of all the incoming ones, except for the one that belongs to the same 
	variable
	\begin{equation*}
		\arraycolsep=1.4pt\def\arraystretch{2.5}
		\begin{array}{llll}
			b_{V^{(i)}}(v) & \propto & \displaystyle \prod_{\mathclap{\substack{j=1 \\ j\neq i}}}^{m}  f_{V^{(j)}}(v) \quad	
			\displaystyle 	\prod_{\mathclap{k=m+1}}^{M}  b_{V^{(k)}}(v), &\; i=1:m \\
			f_{V^{(i)}}(v) & \propto & \displaystyle \prod_{j=1}^{m}  f_{V^{(j)}}(v) \quad \displaystyle 
			\prod_{\mathclap{\substack{k=m+1 \\ k\neq i}}}^{M}  b_{V^{(k)}}(v), &\; i=m+1:M
		\end{array},
	\end{equation*}
	or in vector form
	\begin{equation}
		\arraycolsep=1.4pt\def\arraystretch{2.5}
		\begin{array}{llll} 
			{\bf b}_{V^{(i)}} & \propto & \displaystyle \bigodot_{\mathclap{\substack{j=1 \\ j\neq i}}}^{m}  {\bf f}_{V^{(j)}} 
			\quad \displaystyle 	\bigodot_{\mathclap{k=m+1}}^{M}  {\bf b}_{V^{(k)}}, &\quad i=1:m \\
			{\bf f}_{V^{(i)}} & \propto & \displaystyle \bigodot_{j=1}^{m}  {\bf f}_{V^{(j)}} \quad
			\bigodot_{\mathclap{\substack{k=m+1 \\ k\neq i}}}^{M}  {\bf b}_{V^{(k)}}, &\quad i=m+1:M 
		\end{array}
		\label{eq:prodDiv}
	\end{equation}

	\item The {\em SISO block}, that is the core of the FGrn paradigm, represents the conditional probability matrix $P(Y|V)$ of 
	$Y$ given $V$. Assuming that the output variable $Y$ takes values in the alphabet ${\cal Y}=\{y_{1},...,y_{|{\cal Y}|}\}$, this 
	probability matrix is a $|{\cal V}| \times |{\cal Y}|$ row-stochastic matrix
	\begin{equation*}
		P(Y|V)=[Pr\{ Y= y_{j} | V = v_{i}\}]_{j=1:|{\cal Y}|}^{i=1:|{\cal V}|}=[\theta_{ij}]_{j=1:|{\cal Y}|}^{i=1:|{\cal V}|},
	\end{equation*}
	or more explicitly
	\begin{equation*}
		P(Y|V) =
		\setlength{\arraycolsep}{0.0em}
		\begin{bmatrix} 
			P(Y = y_{1} | V = v_{1}) & \cdots & P(Y = y_{|{\cal Y}|}| V = v_{1}) \\
			P(Y = y_{1} | V = v_{2}) & \cdots & P(Y = y_{|{\cal Y}|}| V = v_{2}) \\
			\vdots & \ddots & \vdots \\
			P(Y = y_{1} | V = v_{|{\cal V}|}) & \cdots & P(Y = y_{|{\cal Y}|}|  V = v_{|{\cal V}|})
		\end{bmatrix} 
		=
		\begin{bmatrix} 
			\theta_{11} & \hdots & \theta_{1|{\cal Y}|} \\
			\theta_{21} & \hdots & \theta_{2|{\cal Y}|} \\
			\vdots & \ddots & \vdots \\
			\theta_{|{\cal V}|1} & \hdots & \theta_{|{\cal V}||{\cal Y}|}
		\end{bmatrix}.
	\end{equation*}
	Outgoing messages are
	\begin{equation*}
		f_Y(y_i) \propto \displaystyle \sum_{j=1}^{|{\cal V}|} \theta_{ij} f_V(v_{j}) \quad \textrm{and} \quad
		b_V(v_j) \propto \displaystyle \sum_{i=1}^{|{\cal Y}|} \theta_{ij} b_Y(y_{i}),
	\end{equation*}
	or in vector form
	\begin{equation}
		{\bf f}_Y \propto P(Y|V)^T {\bf f}_V \quad \textrm{and} \quad
		{\bf b}_V \propto P(Y|V) {\bf b}_Y.
		\label{eq:prodSISO}
	\end{equation}

	\item The {\em Source block} defines an independent  $|{\cal V}|$-dimensional source variable $V$ with its prior 
	distribution $\pi_V$. Therefore, the outgoing message is
	\begin{equation*}
		f_V(v_i)=\pi_V(v_i), \quad i = 1 : |{\cal V}|,
	\end{equation*}
	or in vector form
	\begin{equation*}
		{\bf f}_V={\boldsymbol \pi}_V.
	\end{equation*}
\end{enumerate}
It should be emphasized that the rules presented are a rigorous translation of the total probability theorem and Bayes' rule.

Note  that the only parameters that need to be learned during the training phase are the matrices inside the SISO blocks and the priors inside the Sources. Although different variations are possible \citep{Koller2009, Barber2012, Palmieri2016}, training algorithms are derived mainly as maximization of the likelihood on the observed variables that can be anywhere in the network. Furthermore, within the FGrn paradigm, learning takes place locally; that is, the parameters inside the SISO blocks and the Sources can be learned using only the backward and forward messages to that particular element. This also means that parameter learning in this representation can be addressed in a unified way because we can use a single rule to train any SISO or Source block in the system, simultaneously and independently of its position. Therefore, learning is done iterating over three simple steps:
\begin{enumerate}
	\item present the observations to the network in the form of distributions; they can be anywhere in the network as 
	backward or forward messages; 

	\item propagate the messages in the whole network, in accordance with the mathematical rules just described;

	\item perform the update of SISO blocks and Sources using incoming messages.
\end{enumerate}
The prior of a source is learned simply by calculating the new marginal probability (using \autoref{eq:marginal}), due to the changes of the backward message to the Source. On the other hand, learning the matrices inside the SISO blocks is more complex. According to our experience \citep{Palmieri2016}, the best algorithm for learning the conditional probability matrices inside the SISO blocks is the Maximum Likelihood (ML) algorithm, which uses the following update
\begin{equation} 
	\theta_{l m}^{(1)} \longleftarrow \frac{\theta_{l m}^{(0)}}{\sum_{n=1}^{N} f_{V[n]}(l)}
	\sum_{n=1}^{N} \frac{f_{V[n]}(l) b_{Y[n]}(m)}{{\bf f}_{V[n]}^{T} {\bf \theta}^{(0)} {\bf b}_{Y[n]}}.
	\label{eq:ML}
\end{equation}

\autoref{eq:ML} represents the heart of the ML algorithm and usually requires multiple cycles in order to achieve convergence. However, changing the matrices also changes the propagated messages. For this reason, the whole learning process (starting from the presentation of the evidence) also needs to be performed several times; namely for a fixed number of {\em epochs} (an hyperparameter).
\begin{figure} [!t]
	\centering
	\includegraphics[width=0.4\linewidth]{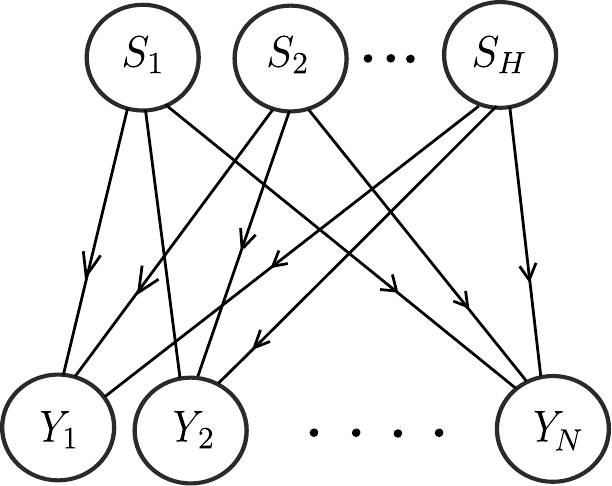}
	\caption[Latent Variable Models represented as Bayesian Graphs]{The general structure for Latent Variable Models}
	\label{fig:LatentVariableModels}
\end{figure}
\noindent

\section{Latent Variable Models} 
\label{sec:LVM}
Factor Graphs and in particular FGrn can be used to represent a  joint probability distribution by appropriately using the dependencies/independence between the variables. In many cases, however, the probabilistic description of an event can be further simplified using a set of unobserved variables, which represent the unknown ``causes" generating the event. In these models the variables are typically divided into {\em Visible} and {\em Hidden}: the first relating to the inputs observed during the inference, the latter belonging to the {\em internal representation}. Obviously, the main advantage of this type of model lies in the fact that it has fewer parameters than the complete model, producing a sort of compressed representation of the observed data (which are therefore easier to manage and analyse).

The simplest and most complete model that includes both visible and hidden variables is the bipartite graph of \autoref{fig:LatentVariableModels}, named Latent Variable Model (LVM) \citep{Murphy2012, Bishop1999} or Autoclass \citep{Cheeseman1996}; any other hidden variables model can ultimately be reduced to such a model. Within the figure, it is possible to distinguish $H$ {\em latent} variables, $ S_1,\hdots,S_H$, and $N$ {\em observed} variables, $ Y_1,\hdots,Y_N$; where typically $N \gg H$. It should be noted that although the given nomenclature seems to subdivide the variables according to their position, where the variables below are known while those above need to be estimated, the bidirectional structure of the network remains quite general, including cases in which some of the above variables may be known and some of the bottom variables need to be estimated. 

\begin{figure} [!t]
	\centering
	\includegraphics[width=0.6\linewidth]{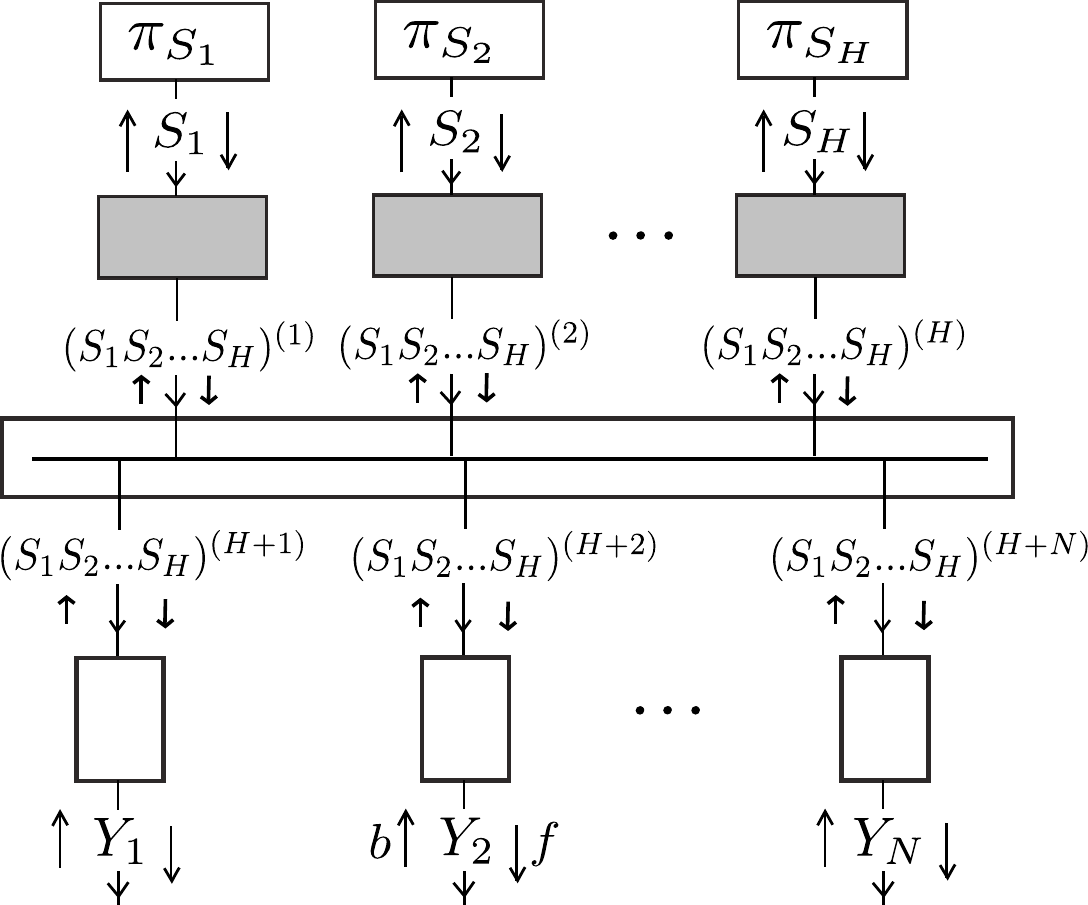}
	\caption[Latent Variable Models represented as Factor Graph in Reduced Normal Form ]{The general structure for the Latent 	
		Variable Models in Reduced Normal Form}
	\label{fig:GeneralNormal}
\end{figure}

The FGrn of a Bayesian network represented by the general LVM structure of \autoref{fig:LatentVariableModels} is shown in \autoref{fig:GeneralNormal}. The system is intrinsically represented as a generative model; that is, choosing to direct the variables downwards and positioning the marginally independent sources $S_1,...,S_H$ at the top. 

Moreover, it should be noted that the supposed independence of all known variables given the hidden variables (provided by the latent variable model) allows to greatly simplify the analysis of the total joint probability, which can now be represented by the factorization
\begin{equation*}
	P(Y_1, \dots ,Y_N, S_1, \dots , S_H)= 
	P(Y_1|S_1, \dots ,S_H) \cdots  P(Y_N|S_1, \dots ,S_H)  P(S_1) \cdots P(S_H) .
\end{equation*}

As said, the structure allows to represent totally heterogeneous variables, but it is good to clarify that (being a representation of a Bayesian network) the variables must be presented as probability vectors; that is, through vectors containing the ``degree of similarity" of the single variable to each particular element of its discrete alphabet. 

The source variables, which have prior distributions $\pi_{S_1}$,~...~$\pi_{S_H}$,  are mapped  to the product space ${\cal P}$, of dimensions $|{\cal P} | = |{\cal S}_1| \times \cdots \times |{\cal S}_H|$, via the fixed row-stochastic matrices (shaded blocks in \autoref{fig:GeneralNormal})
\begin{equation}
	\begin{aligned}
		P((S_1 S_2 \dots S_H)^{(1)}|S_1) &=\frac{|{\cal S}_1|}{\prod_{i=1}^H|{\cal S}_i| }I_{|{\cal S}_1|} 
			\otimes  1_{|{\cal S}_2|}^T \otimes \cdots \otimes  1_{|{\cal S}_H|}^T, \\
		& \vdots \\
		P((S_1 S_2 \dots S_H)^{(H)}|S_H) &=\frac{|{\cal S}_H|}{\prod_{i=1}^H |{\cal S}_i| }1_{|{\cal S}_1|}^T 
			\otimes \cdots \otimes 1_{|{\cal S}_{H-1}|}^T \otimes I_{|{\cal S}_H|} ,
	\end{aligned}
	\label{eq:Mapping}
\end{equation}

\noindent
where $\otimes$ denotes the Kronecker product, $1_K$ is a $K$-dimensional column vector with all ones, and $I_K$ is the $K \times K$ identity matrix \citep{Palmieri2016}. The conditional probability matrix  is such that each variable contributes to the product space with its value, and it is uniform on the components that compete to the other source variables. This is the FGrn counterpart of the Junction Tree reduction procedure because it is equivalent to ``marry the parents'' in Bayesian Graphs \citep{Koller2009}, but here there are explicit branches for the product space variable.

For this reason, although the messages traveling bi-directionally and the initial Bayesian network present many loops (\autoref{fig:LatentVariableModels}), the FGrn architecture will not show any convergence problem because the LVM has been reduced to a tree.

Finally, the $j$-th SISO block at the bottom of \autoref{fig:GeneralNormal}, with $j=1,...,N$, represents the conditional probability matrices $P(Y_j|S_1S_2...S_H)$, which than will have dimensions $|{\cal P}| \times |{\cal Y}_j|$.

\begin{figure} [!t]
	\centering
	\vspace{-1em}
	\includegraphics[width=1.0\linewidth]{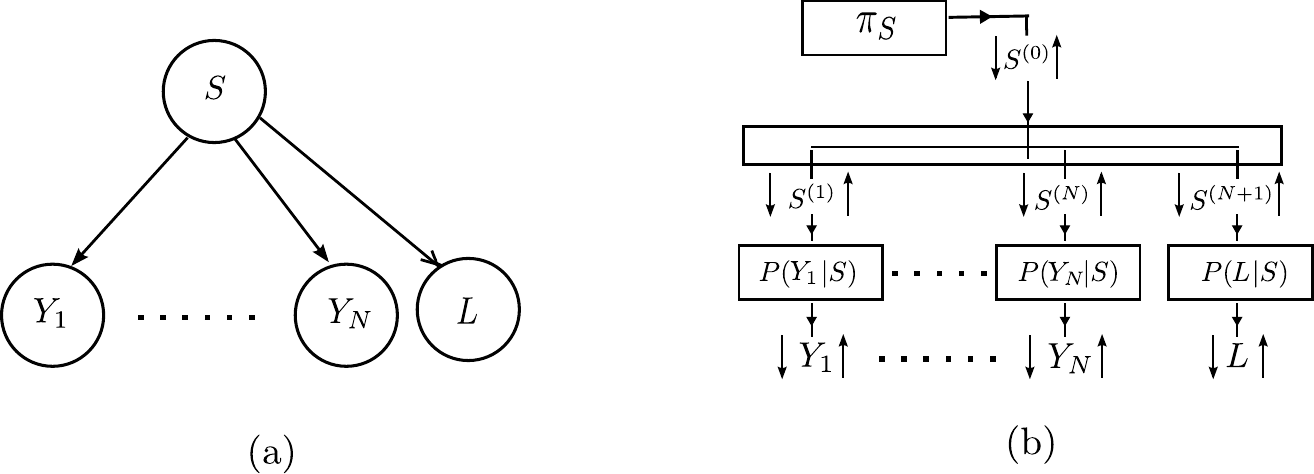}
	\caption{A $N$-tuple with the Latent Variable $S$ and a Class Variable $L$ drawn as (a) Bayesian Network, 
		(b) Factor Graph in Reduced Normal Form.}
	\label{fig:lat1}
	\vspace{-1em}
\end{figure}

\subsection{The LVM with One Hidden Variable}
When $H > 1$ we have a Many-To-Many LVM model, which was already discussed in a previous work \citep{PalmBuon2015}; calling it Discrete Independent Component Analysis (DICA) because it uses the same generative model of the Independent Component Analysis (ICA), but on discrete variables. 

Vice versa, when $H = 1$, we obtain a One-To-Many LVM model, in which there is just one latent factor (parent) that conditions $Y_1,\hdots,Y_N$ (children) and where obviously fixed matrices  (previously represented by the shaded blocks) are no longer necessary. Although the general paradigm has the great advantage of allowing the presence of Sources of different types, for simplicity in the tests performed we have preferred to focus exclusively on this more manageable architecture (\autoref{fig:lat1}). The figure shows the most general case (used in the final tests), in which a Class Variable $L$ is also added to the bottom variables. This configuration is used in the case of supervised learning, allowing (after learning) to perform various tasks, including:
\begin{enumerate}[label=(\alph*)]
	\item {\em Pattern classification}, achievable by injecting the observations as delta distributions on the backwards of 
	$Y_1,...,Y_N$ and leaving the backward on L uniform. The classification will be returned to $L$ through its forward message 
	$f_L$.

	\item {\em Pattern completion}, where only some of the observed variables $Y_1,...,Y_N$ are available in inference (and then 
	injected into the network through delta distributions on their backwards) while the others are unknown (and therefore 
	represented by uniform backward distributions). 
	
	Also $L$ may be unknown (uniform distribution), partially known (generic distribution), or 
	known (delta distribution). The forward distributions at the unknown variables complete the pattern, and at $L$ provides the best 
	inference in case it is not perfectly known. The posterior on $L$ is obtained by multiplying the two messages on it 
	(\autoref{eq:marginal}); avoidable step if $L$ is not known at the beginning (because in this case the backward is 
	uniform).

	\item {\em Prototype inspection}, obtainable by injecting only a delta distribution at $L$ on the $j$th label. The forward 
	distributions $f_{Y_1},...,f_{Y_N}$ will represent the separable prototype distribution of that class. 
\end{enumerate}
Another way to use the previous network is to make the class variable coincide with the hidden variable $S=L$, forcing the corresponding SISO block matrix to be diagonal. This constraint will create a so-called ``Naive Bayes Classifier'' \citep{Barber2012}, further simplifying the factorization into
\begin{equation}
	P(Y_1, ... ,Y_N, L)=P(Y_1|L) \cdots  P(Y_N|L)  P(L). 
\end{equation}
In this case, usually all the variables are observed during training, and the typical use in inference is to obtain $L$ from observed $Y_1,...,Y_N$ .

Note that the case related to unsupervised learning can be obtained from the general model presented simply by eliminating the variable $L$. In this case, after learning, the elements of the alphabet ${\cal S}=\{ s_1,...,s_{|{\cal S}|}\}$ of the hidden variable $S$ represent  ``Bayesian clusters" of the data, which follows the  prior distribution $\pi_S$ (learned blindly). The network can be used in inference both for the pattern completion, in the case where (as seen previously) only some of the underlying variables are known and we try to estimate the others through the corresponding forward messages, and to create a so-called embedding representation, in which the backward message becomes a different representation of the underlying known variables. In the latter case, in order to understand the representation that the network has created, we can look at the j-th centroid of the bayesian clusters injecting as ${\bf f}_{S}$ a delta distribution ${\bf \delta}_j = [0 \hdots 1 \hdots 0]^T$, where the $1$ is at the j-th position. The set of forward distributions $f_{Y_1},...,f_{Y_N}$ generated by the network will represent the marginal distributions around the centroid of the $j$th cluster. 

\section{Design of FGrnLib}
\label{sec:library}
There are several software packages, known in the literature, that can be used to design and/or  simulate Bayesian networks (an updated list can be found in \citep{MurphySWPackage}). Unfortunately, many of them are in closed packages  and/or run only on private servers, preventing proper performance analysis.  Others either have limitations on the number of variables and the size of the network, or do not use the FG architecture. Therefore, the main purpose of this work was to design an optimized library, called {\em FGrnLib}, for the realization of a Bayesian network through the use of the FGrn model; that is open and contains an efficient implementation of the elements in \autoref{fig:blocks}.

The {\em FGrnLib} library has been written in C++, following the classic object-oriented paradigm, and it has been adapted for parallel computing (on multiprocessor systems with shared memory) through the use of the OpenMP application interface. 

The various algorithmic operations  have been implemented to limit as much as possible the computational complexity without significantly affecting memory requirements.

\begin{figure*} [!t]
	\centering
	\vspace{-1em}
	\includegraphics[width=1.0\linewidth]{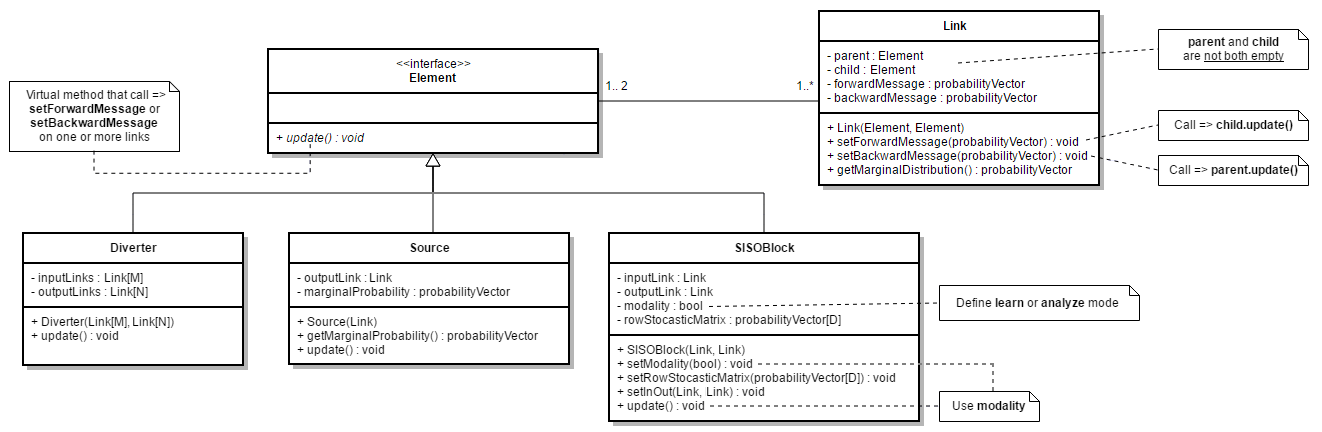}
	\caption{The basic class diagram for FGrnLib, that show the dependencies between the various classes.}
	\label{fig:classes}
	\vspace{-1em}
\end{figure*}

\vspace{-0.5em}
\subsection{Data Structures}
Before starting the analysis of individual operations, it is necessary to focus on the structure of the main classes (\autoref{fig:classes}) and their individual roles. These classes correspond to the main elements presented in \autoref{fig:blocks} (and mathematically described in \autoref{sec:rnf}), but also have subtle design choices that need to be clarified (especially in reference to what was previously done in \citep{BuonPalm2015}). The main classes are:
\begin{itemize}
	\item The \texttt{Link} class, which represents a single discrete variable of the model. This class contains the two 	
	forward and backward messages for the variable, and is designed to ensure that each message at each time 
	step is represented by a single vector. This means that in every instant of time every single link takes on only two 
	messages (in the two different directions); providing better control of the information traveling on the network but preventing the 
	possibility of carrying out stages of learning through the simultaneous presentation of all the evidence.

	\item The \texttt{Diverter} class, which imposes the constraint of equality among the variables. This class has been created to be 	
	as general as possible, in the sense that it can automatically adapt the parameters of the net to the number of variables. For this 
	reason, it includes not only the process of replication of variables but also the creation and control of product space matrices 
	(\autoref{eq:Mapping}). For space complexity, these (sparse and row-stochastic) matrices are stored in column vectors, whose 
	elements represent the index of the active column in that particular row of the matrix
	\begin{equation*}
		\begin{bmatrix} 1 & 0 & 0 \\ 0 & 1 & 0 \\ 0 & 0 & 1 \\ 1 & 0 & 0 \\[-0.5em] & \vdots &  \end{bmatrix}
		\longrightarrow
		\begin{bmatrix} 0\\1\\2\\0\\[-0.5em] \vdots \end{bmatrix}.
	\end{equation*}.

	\item The \texttt{SISOBlock} class, which represents the probability of the output variable $Y$ given an input variable $V$, contains the row-stochastic conditional probability matrix $P(Y|V)$. Due to the fact that the \texttt{Link} class allows 
	only two vector at a time, the SISO blocks are realized to permit the storage and retrieval of all messages that reached the blocks 
	during the batch learning phase; avoiding transmitting the evidences all at the same time but reusing the single \texttt{Link} 
	memory units for each step.

	\item The \texttt{Source} class, which represents the independent variables inside the model, with their prior probabilities.
\end{itemize}

\begin{figure} [ht]
	\centering
	\vspace{-1em}
	\includegraphics[width=1.0\linewidth]{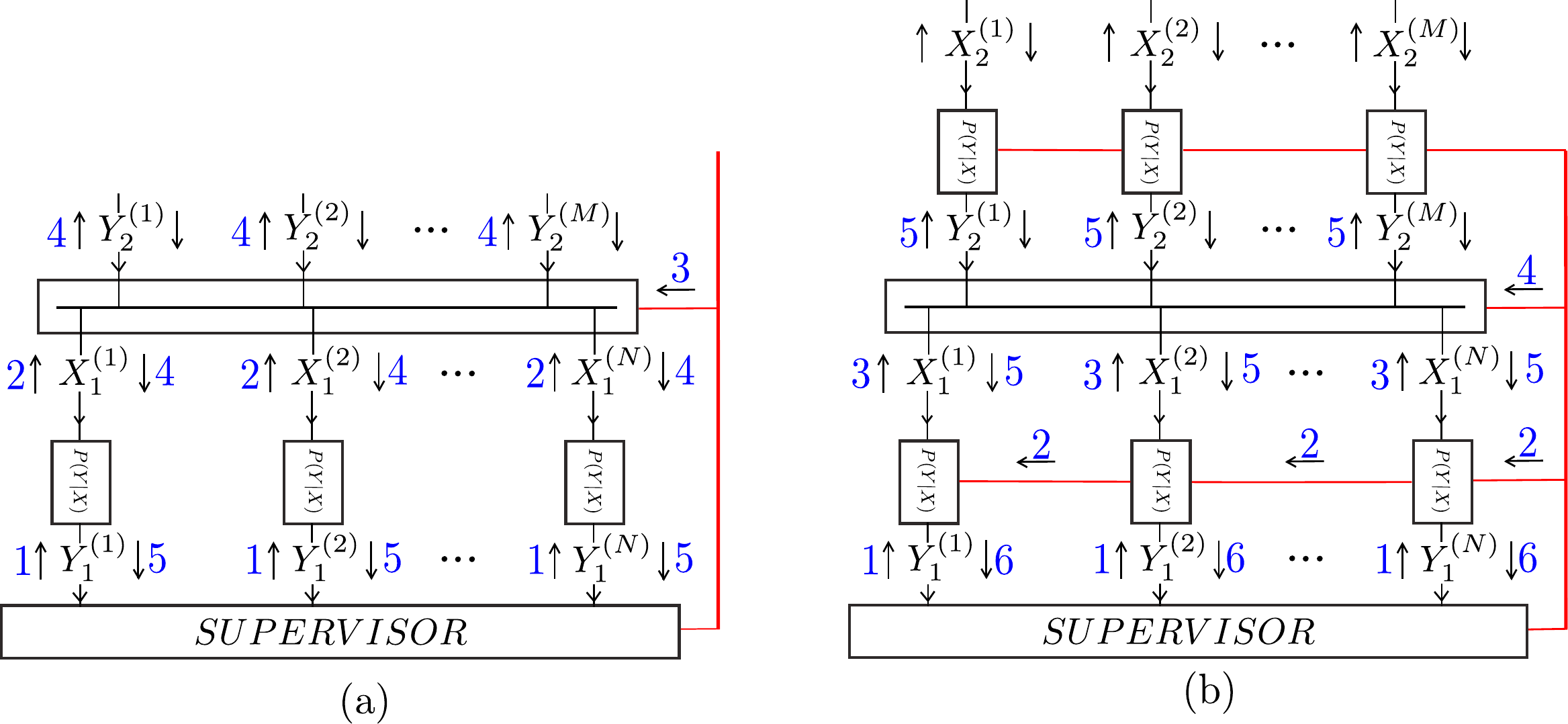}
	\vspace{-1em}
	\caption{ Messages propagation: (a) inference phase; (b) batch learning phase.}
	\label{fig:FlowExample}
\end{figure}

\subsection{Data Flow Management}
Since the FGrn paradigm allows to work only through local information, the flow of messages within the network can be parallelized. However, parallelizing the message flow in the network imposes essential changes to the Diverter class.  In fact, the multiplication of the messages internally to the Diverter can take place only after all the messages of the variables relating thereto have been updated. Being the only responsible for the combination of messages coming from different directions, the Diverter must necessarily act as a ``barrier" (with the meaning that this term assumes in parallel programming); solving the synchronization problems simply by not updating the output values until it receives the activation signal from a {\em supervisor}. Although within the FGrnLib a \texttt{Supervisor} class has been defined to more easily manage some predefined network, the meaning of a supervisor is quite general here because it refers to any class that possesses all the references to the single elements of the network realized.

In \autoref{fig:FlowExample} it is shown how the supervisor handles the message scheduling, relatively to both the inference mode and the batch learning mode. Note that, in order for messages that travel in parallel to be propagated anywhere on the network, a number of steps equal to the diameter of the graph is required \citep{Pearl1988}. 

Recalling therefore that, in graph theory, the diameter of the network  is the longest possible path (in terms of number of crossed arcs) among the smallest existing paths able to connect each vertex inside the graph, it is easy to understand that a simple one-layer LVM network (like in \autoref{fig:GeneralNormal} or in \autoref{fig:lat1}.a) will only need three propagation steps. 

\subsubsection{Inference}
The inference phase, depicted in \autoref{fig:FlowExample}.a, is relatively simple: the various messages proceed in parallel until they reach the Diverter. All that is necessary is to block the messages at the Diverter, to prevent the start of the multiplication process before all the messages have become available. The supervisor must therefore perform only two phases: parallelize the input variables and then activate the Diverter. It should be noted that typically all messages are initialized to uniform vectors, and updated according to the rules described in \autoref{sec:rnf}. 

For what has been said, in the first step of \autoref{fig:FlowExample}.a the supervisor performs the parallelization of the input variables $Y_1, \dots ,Y_N$, placing in the backward messages appropriate distributions for the known (or partially known) variables and uniform distributions for the unknown ones. Hence, in the second step, the messages start to be propagated in the network (using the second in \autoref{eq:prodSISO}). At this point, when all the variables have performed the second step, the supervisor must activate the Diverter to allow the propagation of the messages to continue. The last message propagation step allows to obtain the desired output values (using the first in \autoref{eq:prodSISO}).

\subsubsection{Batch Learning} 
Every single epoch of the batch learning phase (\autoref{fig:FlowExample}.b) is not so different from what we have just seen for the inference phase, since the supervisor basically performs only two more operations. The first one, which only applies at the beginning of the whole learning phase (that is, it is not reiterated at each epoch), consists in activating the batch learning mode inside the SISO blocks and the Sources. This step enables storage within the SISO blocks, which consequently memorize the incoming vectors (from both sides) as soon as they are updated, and the Sources, which will begin to add together all incoming backward messages within a temporary vector. It should be noted that, in the case of multi-layer structures, the supervisor will also have to worry about preventing the SISO blocks of the next level from propagating incoming messages, thus transforming the latter into ``barriers" as well. In fact, in order to get the classic layer-by-layer approach, the information should not propagate above the Diverter until the underlying batch learning phase is complete, but only if the top layer does not provide Sources. So as not to complicate the Diverter too much, forcing it to know the connected elements, the SISO blocks also provide a pause mode, which if enabled prevents forward propagation of messages.

After activation of the batch learning mode the messages propagate within the network through the same previous steps, being blocked again to the Diverter. However, as we can see in \autoref{fig:FlowExample}.b, the descending phase will not include the production of the final messages, which will only be stored in the SISO blocks in order to avoid performing the related mathematical operations.

The second different operation, performed by the supervisor at the end of each epoch, consists in the activation of the actual learning procedure, which will execute the ML algorithm using the stored vectors (up to a sufficient convergence or at most for a fixed number of times) and change the prior of the Source. After this last phase, the procedure can be repeated for another epoch, presenting the evidence to the network again. When it will be necessary to end the learning, the last message of the supervisor will also modify the functioning modalities of the SISO blocks and the Sources, thus making them ready for the subsequent phases of inference.

\section{Complexity and Efficient Algorithms}
\label{sec:complexity}
Having to pay particular attention to the computational and memory costs, in the creation of the library we worked on the details of each individual element. This, together with the probabilistic nature of the Bayesian networks, has led to the preliminary definition of particular basic data structures that it is perhaps necessary to analyse quickly. In fact, the library also defines classes that represent probability vectors and row-stochastic matrices, to facilitate the interpretation and definition of the variables and to easily manage all the algebraic operations.

\subsection{Probability Vector}
\label{sub:vector}
A probability vector is a vector with elements in $[0,1]$ that sum to one. Although the network does not use only probability vectors, the execution of every operation that uses them must then provide normalization. Every normalization consists of $d - 1$ sums and $d$ divisions, so the computational cost will  be $\mathcal{O}(d)$
\begin{equation*}
	\left.
	\begin{bmatrix} \upsilon_1\\ \upsilon_2 \\ \vdots \\ \upsilon_d \end{bmatrix} \right |_{norm}
	= 
	\begin{bmatrix} \zeta_1\\ \zeta_2 \\ \vdots \\ \zeta_d \end{bmatrix}
	\longrightarrow
	\zeta_i = \frac{\upsilon_i}{\sum_{j=1}^d \upsilon_j}.
\end{equation*}

\subsection{Row-stochastic Matrix Multiplication}
In a row-stochastic matrix $P$ each row is a probability vector. It is important to observe that the premultiplication or postmultiplication of a row-stochastic matrix for a vector (with the appropriate dimensions)
\begin{equation*}
	\setlength{\arraycolsep}{0.2em}
	\begin{aligned} 
		\begin{bmatrix} \xi_1 & \hdots & \xi_l \end{bmatrix}
		\begin{bmatrix} p_{1 1} & \cdots & p_{1 d} \\ \vdots & \ddots & \vdots \\  p_{l 1} & \cdots &  p_{l d} \end{bmatrix} 
		& = 
		\begin{bmatrix} \zeta_1 & \hdots & \zeta_l \end{bmatrix}
		& \longrightarrow
		\zeta_j = \sum_{i=1}^l p_{ij} \xi_i \\[1em]
		\begin{bmatrix} p_{1 1} & \cdots & p_{1 d} \\ \vdots & \ddots & \vdots \\  p_{l 1} & \cdots &  p_{l d} \end{bmatrix} 
		\begin{bmatrix} \xi_1\\ \vdots \\ \xi_d \end{bmatrix}
		& = 
		\begin{bmatrix} \zeta_1\\ \vdots \\ \zeta_l \end{bmatrix}
		& \longrightarrow
		\zeta_i = \sum_{j=1}^d p_{ij} \xi_i
	\end{aligned} 
\end{equation*}
will consist of $ld$ multiplications and $(l-1)d$ or $l(d - 1)$ sums respectively, producing the same computational cost $\mathcal{O}(ld)$.

\subsection {Diverter}
From a computational point of view, the most critical structure in the implementation of a Bayesian network using FGrn is represented by the Diverter, where, obviously, the greatest criticality is in the efficient implementation of the internal multiplication process (\autoref{eq:prodDiv}). 

\begin{figure}[ht]
	\centerline{\includegraphics[width=0.4\textwidth]{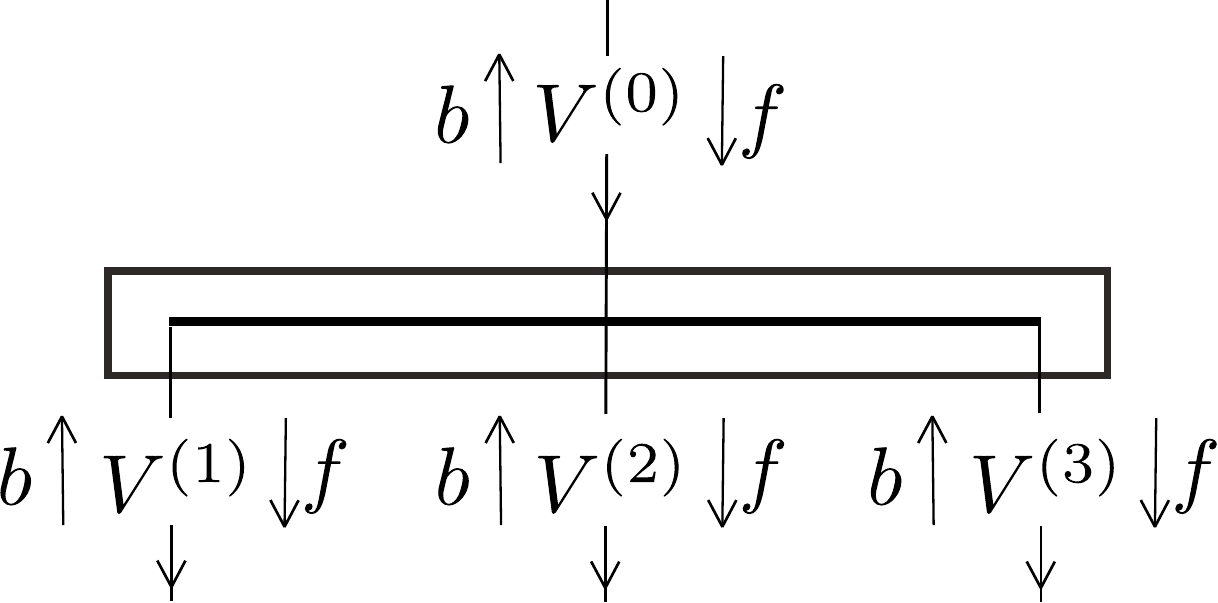}}
	\caption{A diverter with $M=4$ (one input and three output variables).}
	\label{fig:Example1}
\end{figure}

\begin{figure*} [!t]
	\centering
	\includegraphics[width=0.9\linewidth]{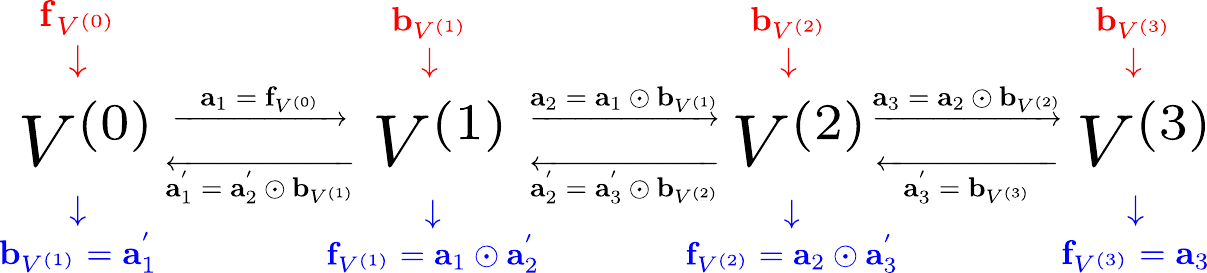}
	\caption{Details of the efficient implementation of the products inside the Diverter, with in red the {\em input messages} and in 
		    blue the {\em output messages}. For each output message two contributions are used: one derived from the left part of 
		    the computational graph and the other one derived from the right part.}
	\label{fig:DiverterComplexity}
\end{figure*}

In fact, the possible presence of zeros, in different positions of the single vectors, obliges to perform the calculation of the outgoing vectors individually. In the general case of \autoref{fig:GeneralNormal}, indicating the total value of all the variables connected to the diverter with $M = H + N$ and assuming that the input vectors have dimensions equal to $d$, bare application of the multiplication rule would require $M (M - 2) d$ multiplications. Regardless of the size of the input vectors, the computational cost is therefore polynomial and equal to $\mathcal{O}(M^{2})$ vectorial multiplications. Considering as an example the Diverter of \autoref{fig:Example1} (with $M=4$), the simple application of the product rule in the production of the outgoing messages $({\bf b}_{V^{(0)}}, {\bf f}_{V^{(1)}}, {\bf f}_{V^{(2)}}, {\bf f}_{V^{(3)}})$ would requires eight vectorial multiplications:
\begin{equation*}
	\begin{aligned} 
		{\bf b}_{V^{(0)}}&={\bf b}_{V^{(1)}} \odot {\bf b}_{V^{(2)}} \odot {\bf b}_{V^{(3)}}; \\
	     	{\bf f}_{V^{(1)}}&={\bf f}_{V^{(0)}} \odot {\bf b}_{V^{(2)}} \odot {\bf b}_{V^{(3)}}; \\
		{\bf f}_{V^{(2)}}&={\bf f}_{V^{(0)}} \odot {\bf b}_{V^{(1)}} \odot {\bf b}_{V^{(3)}}; \\
	     	{\bf f}_{V^{(3)}}&={\bf f}_{V^{(0)}} \odot {\bf b}_{V^{(1)}} \odot {\bf b}_{V^{(2)}}.
	\end{aligned} 
\end{equation*}

This process can be performed more efficiently by defining an order among the variables connected to the Diverter and by performing a double cascade process in which each variable is responsible only for passing the correct value to the neighboring variable. In this way, the variables at the ends of the chain will  perform no multiplication while each variable inside the chain will perform only three multiplications, relative to the passage of the two temporary vectors along the chain and to the output of the outgoing message. 

With reference to the example of \autoref{fig:Example1} we have the data flow represented in the \autoref{fig:DiverterComplexity}. In other words, the proposed solution exploits the presence of the same multiplication groups through a round-trip process. This reduces the computational complexity from quadratic to linear, $\mathcal{O}(M)$, finally requiring only $3(M - 2)$ vector multiplications. Although there is obviously an increase in the memory required, due to temporary vectors along the chain, it remains linear with $M$, and has been further optimized (requiring only $M - 1$ temporary vectors altogether) by choosing to reuse the same vectors ($a_i=a_i^{'}$) when changing direction.

\subsection{Unknown variables}
A very attractive property of using a probabilistic paradigm, which makes it preferable in certain contexts, is represented by its ability to manage unknown inputs. 

Even in the case of maximum uncertainty, that is when nothing is known about the particular variable in that particular observation set, a process of inference or learning can still be performed by making the corresponding message of that variable a uniformly distributed probability vector
\begin{equation*}
	\bar{\bf b}_{Y_i}=\begin{bmatrix} \frac{1}{|{\cal Y}_i|} \\ \vdots \\ \frac{1}{|{\cal Y}_i|} \end{bmatrix}.
\end{equation*}
In this particular circumstance, the message propagation process can be optimized by avoiding the multiplication of backward vectors with the matrices inside the SISO blocks, noting that
\begin{equation*}
	\setlength{\arraycolsep}{0.0em}
	\begin{aligned}
		{\bf b}_{S^{(i)}} = P(Y_i | S)\bar{\bf b}_{Y_i}
		&= 	\begin{bmatrix} 
				\scalebox{0.9}{$P(Y_i = \xi_1| S = \sigma_1)$} & \cdots &  
				\scalebox{0.9}{$P(Y_i = \xi_{|{\cal Y}_i|}| S = \sigma_1)$} \\ 
				\vdots & \ddots & \vdots \\ 
				\scalebox{0.9}{$P(Y_i = \xi_1| S = \sigma_{|{\cal S}|})$} & \cdots & 
				\scalebox{0.9}{$P(Y_i = \xi_{|{\cal Y}_i|}| S = \sigma_{|{\cal S}|})$}
			\end{bmatrix}
			\begin{bmatrix}\frac{1}{|{\cal Y}_i|} \\ \vdots \\ \frac{1}{|{\cal Y}_i|} \end{bmatrix} \\
		&=	\begin{bmatrix} 	
				\displaystyle \sum_{j=1}^{|{\cal Y}_i|} \theta_{1 j} {\bar b}_{Y_i} (\xi_j) \\ \vdots \\
				\displaystyle \sum_{j=1}^{|{\cal Y}_i|} \theta_{|{\cal S}| j} {\bar b}_{Y_i} (\xi_j) 
			\end{bmatrix}
		  =	\begin{bmatrix} 	
				\frac{1}{|{\cal Y}_i|} \displaystyle \sum_{j=1}^{|{\cal Y}_i|} \theta_{1 j} \\ \vdots \\
				\frac{1}{|{\cal Y}_i|} \displaystyle \sum_{j=1}^{|{\cal Y}_i|} \theta_{|{\cal S}| j} 
			\end{bmatrix}
		  =	\begin{bmatrix}\frac{1}{|{\cal Y}_i|} \\ \vdots \\ \frac{1}{|{\cal Y}_i|} \end{bmatrix}.
	\end{aligned}
\end{equation*}

By not propagating the unknown variable (setting ${\bf b}_{S^{(i)}}$ as an expanded/reduced version of $\bar{\bf b}_{Y_i}$), for every single unknown variable present in input during the inferential and the learning process we can save $|{\cal S}|$ vector multiplications, improving overall network performance.

\subsection{Efficient ML Implementation}
Regarding the learning phase, particular attention has been given to the realization of an efficient implementation of the ML algorithm. First of all, since the matrix inside the SISO blocks is set to be row-stochastic by construction, it has been noted that the first divisor in the \autoref{eq:ML} becomes unnecessary. For this reason, the equation can be rewritten as follows
\begin{equation*} \label{eq:ML2}
	\theta_{l m}^{(1)} \longleftarrow \theta_{l m}^{(0)}
	\sum_{n=1}^{N} \frac{f_{V[n]}(l) b_{Y[n]}(m)}{{\bf f}_{V[n]}^{T} {\bf \theta}^{(0)} {\bf b}_{Y[n]}},
\end{equation*}
or in vector form
\begin{equation*}
	{\bf \theta}^{(1)} \longleftarrow {\bf \theta}^{(0)} \odot 
	\sum_{n=1}^{N} \frac{{\bf f}_{V[n]} {\bf b}_{Y[n]}^{T}}{{\bf f}_{V[n]}^{T} {\bf \theta}^{(0)} {\bf b}_{Y[n]}}.
\end{equation*}

Furthermore, it can be observed that the value obtained through the vector multiplications ${\bf f}_{V[n]}^{T} {\bf \theta}^{(0)} {\bf b}_{Y[n]}$ is actually equal to the sum of all elements of the matrix ${\bf \theta}^{0} \odot {\bf f}_{V[n]} {\bf b}_{Y[n]}^{T}$. 
This assertion is provable by observing that
\begin{equation*}
	\begin{aligned}
		{\bf \theta}^{0} \odot {\bf f}_{V[n]} {\bf b}_{Y[n]}^{T}
		& =	\begin{bmatrix} 
				\theta_{1 1}^{(0)} & \cdots &  \theta_{1 |{\cal Y}|}^{(0)} \\ 
				\vdots & \ddots & \vdots \\ 
				\theta_{|{\cal V}| 1}^{(0)} & \cdots &  \theta_{|{\cal V}| |{\cal Y}|}^{(0)}
			\end{bmatrix} 
		      	\odot 
			\begin{bmatrix} \phi_{1} \\ \vdots \\ \phi_{|{\cal V}|} \end{bmatrix}
			\begin{bmatrix} \beta_{1} & \cdots & \beta_{|{\cal Y}|} \end{bmatrix} \\
		& =	\begin{bmatrix} 
				\theta_{1 1}^{(0)} & \cdots &  \theta_{1 |{\cal Y}|}^{(0)} \\ 
				\vdots & \ddots & \vdots \\ 
				\theta_{|{\cal V}| 1}^{(0)} & \cdots &  \theta_{|{\cal V}| |{\cal Y}|}^{(0)} 
			\end{bmatrix} 
		      	\odot 
			\begin{bmatrix} 
				\phi_{1}\beta_{1} & \cdots &  \phi_{1}\beta_{|{\cal Y}|} \\ 
				\vdots & \ddots & \vdots \\ 
				\phi_{|{\cal V}|}\beta_{1} & \cdots &  \phi_{|{\cal V}|}\beta_{|{\cal Y}|} 
			\end{bmatrix} \\
		 & =	\begin{bmatrix} 
				\theta_{1 1}^{(0)}\phi_{1} \beta_{1} & \cdots &  \theta_{1 |{\cal Y}|}^{(0)} \phi_{1} \beta_{|{\cal Y}|} \\ 
				\vdots & \ddots & \vdots \\ 
				\theta_{|{\cal V}|1}^{(0)} \phi_{|{\cal V}|} \beta_{1} & \cdots 
				&  \theta_{|{\cal V}| |{\cal Y}|}^{(0)} \phi_{|{\cal V}|}\beta_{|{\cal Y}|}
			\end{bmatrix},
	\end{aligned}
\end{equation*}
and realizing that the sum of all the elements of the matrix can be written in the form
\begin{equation*}
	\sum^{|{\cal V}|}_{l=1} \sum_{m=1}^{|{\cal Y}|} {\bf \theta}^{(0)} \odot {\bf f}_{V[n]} {\bf b}_{Y[n]}^{T} = 
	\sum_{l=1}^{|{\cal V}|} \phi_{l} \sum_{m=1}^{|{\cal Y}|} \theta_{l m}^{(0)} \beta_{m},
\end{equation*}
which precisely is equal to
\begin{equation*}
	\begin{split}
		{\bf f}_{V[n]}^{T} {\bf \theta}^{(0)} {\bf b}_{Y[n]}
		&= 	\begin{bmatrix} \phi_{1} & \cdots & \phi_{|{\cal V}|} \end{bmatrix}
			\begin{bmatrix} 
				\theta_{1 1}^{(0)} & \cdots &  \theta_{1 |{\cal Y}|}^{(0)} \\ 
				\vdots & \ddots & \vdots \\ 
				\theta_{|{\cal V}| 1}^{(0)} & \cdots &  \theta_{|{\cal V}| |{\cal Y}|}^{(0)} 
			\end{bmatrix} 
			\begin{bmatrix} \beta_{1} \\ \vdots \\ \beta_{|{\cal Y}|} \end{bmatrix} \\
		&=	\begin{bmatrix} \phi_{1} & \cdots & \phi_{|{\cal V}|} \end{bmatrix}
			\begin{bmatrix} 	
				\sum_{m=1}^{|{\cal Y}|} \theta_{1 m}^{(0)} \beta_{m} \\ \vdots \\
				\sum_{m=1}^{|{\cal Y}|} \theta_{|{\cal V}| m}^{(0)} \beta_{m} 
			\end{bmatrix}
		  =	\sum_{l=1}^{|{\cal V}|} \phi_{l} \sum_{m=1}^{|{\cal Y}|} \theta_{l m}^{(0)} \beta_{m}.
	\end{split}
\end{equation*}
This suggests that moving the Hadamard product inside the summation 
\begin{equation*}
	{\bf \theta}^{(1)} \longleftarrow \sum_{n=1}^{N} 
	\frac{{\bf \theta}^{(0)} \odot {\bf f}_{V[n]} {\bf b}_{Y[n]}^{T}}{{\bf f}_{V[n]}^{T} {\bf \theta}^{(0)} {\bf b}_{Y[n]}}
\end{equation*} 
and calculating the sum of all the elements of the matrix ${\bf \theta}^{0} \odot {\bf f}_{V[n]} {\bf b}_{Y[n]}^{T}$  at the same time of their generation, computational complexity of the algorithm can overall be reduced from $N(3 |{\cal Y}| + 1) |{\cal V}|$ to $2 N |{\cal V}| |{\cal Y}|$ multiplications, which is rather significant if considering that this reduction is relative to each algorithm recall.

\section{Performance on LVM}
\label{sec:performance}
To evaluate the computational advantages obtained by the proposed improvements let us consider the more general situation of \autoref{fig:GeneralNormal}, in which the LVM (single-layer) model foresees $H$ sources (hidden) and $N$ output (observed) variables $Y_1,...,Y_N$. Since the variables $Y_1,...,Y_N$ can have different dimensions (never less than $2$), before starting it is important to underline that  we will assume that all the observed variables have the same dimension $|{\cal Y}|$. This statement does not compromise in any way the goodness of the results obtained, for we can certainly decide to choose $|{\cal Y}|$ as the highest value possible being interested only in an upper-bound description of computational complexity (in terms of big-O notation). For the same reason, we will avoid considering the case (more advantageous) in which some variables are not know. Note that, the previous problem does not arise in the case of input variables to the Diverter because they already have the same dimension $|{\cal P}|$ by construction.

\subsection{Cost of the inference phase}
In the inference mode, at the beginning of the process, the backward messages of the $N$ output variables will be post-multiplied by the probabilistic matrix inside the SISO blocks; thus producing $\mathcal{O}(N|{\cal P}||{\cal Y}|)$ operations before being sent to the Diverter. As seen previously, $\mathcal{O}((H+N)|{\cal P}|)$ operations will be performed inside the Diverter, related to the multiplication of all the $H+N$ incoming messages to it. Finally, the messages must be propagated again to the SISO blocks, which will be pre-multiplied by the matrices still producing $\mathcal{O}(N|{\cal P}||{\cal Y}|)$ operations, and thus making the total computational cost equal to $\mathcal{O}((N|{\cal Y}|+H)|{\cal P}|)$. At this point the forward messages of the $Y_1, \dots, Y_N$ output variables will be available for analysis. The following table shows the difference in computational terms determined by the introduced optimizations

\begin{center}
	\begin{tabular}{r|c|}
		\cline{2-2}
		\rule[-0.25ex]{0pt}{3ex} & Computational Cost\\ \cline{2-2}
		\rule[-0.25ex]{0pt}{3ex} Direct & $\mathcal{O}((N(|{\cal Y}|+H+N)+H^2)|{\cal P}|)$ \\ \cline{2-2}
		\rule[-0.25ex]{0pt}{3ex} Optimized & $\mathcal{O}((N|{\cal Y}|+H)|{\cal P}|)$ \\ \cline{2-2}
	\end{tabular}
\end{center}
\vspace{1em}

To provide a practical example, suppose to perform an inferential process on an LVM network with $10$ binary output variables and a single hidden variable, with an embedding space $|{\cal S}|$ of $10$. It will thus be noted that while the direct algorithm will require 1500 multiplications, the optimized one will require only 670. 

Moreover, the additional memory required by the optimized algorithm is still linear with the size of the inputs, since it depends only on the temporary vectors inside the Diverter. In fact, in the previous example, the significant computational advantage obtained will correspond to an increase in memory equal to only $10$ vectors of size $10$.

\subsection{Cost of the batch learning phase}
For what concerns the batch learning session, assume to have $L$ different examples given in input through the backward messages of the output variables. As already mentioned, when the process begins backward messages entering the network will be saved inside the SISO blocks, requesting an amount of memory of $\mathcal{O}(L|{\cal Y}|)$ individually. After storing the vector, the process will continue by multiplying it and the probability matrix; sending the result to the Diverter. In the propagation phase the computational costs will not change with respect to what has been seen in the inference phases, but it must be remembered that the messages that will return back to the SISO blocks will be memorized again; with memory costs individually equal to $\mathcal{O}(L|{\cal P}|)$. The total cost of additional memory required is therefore $\mathcal{O}(LN(|{\cal Y}|+|{\cal P}|))$, while the computational cost is still $\mathcal{O}(L(N|{\cal Y}|+H)|{\cal P}|)$.

Once this first phase has been completed, the ML algorithm will be executed at most K times (with K fixed a priori), trying to make conditional probability matrices converge, and the whole process is then repeated $T$ times. Thus, the total computational cost of the batch learning session is equal to 
\begin{equation*}
	\mathcal{O}(TL(KN|{\cal Y}|+H)|{\cal P}|).
\end{equation*}
Again, a table is presented which facilitates the comparison between the computational costs of the direct case and those of the optimized case.

\begin{center}
	\begin{tabular}{r|c|}
		\cline{2-2}
		\rule[-0.25ex]{0pt}{3ex} & Computational Cost \\ \cline{2-2}
		\rule[-0.25ex]{0pt}{3ex} Direct & $\mathcal{O}(TL(N(K|{\cal Y}|+H+N)+H^2)|{\cal P}|)$ \\ \cline{2-2}
		\rule[-0.25ex]{0pt}{3ex} Optimized & $\mathcal{O}(TL(KN|{\cal Y}|+H)|{\cal P}|)$ \\ \cline{2-2}
	\end{tabular}
\end{center}
\vspace{1em}

Furthermore, it is easy to state that the repetition of the process for $T$ epochs, as well as the various calls of the ML algorithm, do not imply the need for any additional memory units with respect to the non-optimized case.

\section{Incremental Algorithm}
\label{sec:incremental}
As noted above, the ML algorithm has many undoubted advantages, being very stable and typically converging in a few steps. Unfortunately it is a batch-type algorithm that is able to perform only on the whole training set. In order to obtain a lighter implementation we have changed  the previous structure by requiring that at each epoch of the learning phase  only one ML cycle (that is, $K=1$) is included. In other words, the algorithm has been made incremental obtaining the advantage of reducing the amount of memory. In fact, this approach makes it unnecessary to store backward messages within SISO blocks, since they must now be used only once. This eliminates the need for the previous storage space equal to $L(|{\cal P}|+|{\cal Y}|)$ for each SISO block.

Despite the great advantage in terms of both memory and computational costs, this type of approach has surprisingly proved to be as robust as the previous one, being less likely to provide overfitting of the data. Referring to a One-To-Many LVM structure (\autoref{fig:lat1}.b), in which the only hidden variable has an embedding space $|{\cal S}|$ of $20$, the following tables show the accuracy of the classification for three different datasets.

\begin{center}
	\begin{threeparttable}
		\begin{tabular}{r|c|c|c|c|}
			\cline{2-5}
			& \multicolumn{2}{|c|}{Training set} 
			& \multicolumn{2}{|c|}{Test set}\\ \cline{2-5}
			& Batch & Incremental & Batch & Incremental\\ \cline{2-5}
			Breast Cancer & 96,6\% & 96,2\%& 95,48\% & 97,99\%\\ \cline{2-5}
			Mammographic Mass & 86,62\% & 85,88\%& 79,5\% & 78,88\%\\ \cline{2-5}
			Contraceptive Method & 58,1\% & 58,9\%& 49,89\% & 50,52\%\\ \cline{2-5}
		\end{tabular}
		\begin{tablenotes} \small \vspace{0.2em}
	      		Datasets composition: Breast Cancer = 10 useful variables, 699 instances, 16 missing values; 
			Mammographic Mass = 6 variables, 961 instances, 162 missing values; 
			Contraceptive Method = 10 variables, 1473 instances, 0 missing values.
	    	\end{tablenotes}
	\end{threeparttable}
\end{center}

The tables present the classification success rates, both on the training and on the test set, of three databases from the UCI repository: Wisconsin Breast Cancer Dataset \citep{Dua2017}, Mammographic Mass Data Set \citep{Elter2007} and Contraceptive Method Choice Data Set \citep{Dua2017}. The values presented are obtained by making sure that the latent variable is not learned, and therefore represent the learning ability of a single layer; being a good index to represent layer-by-layer learning of more complex networks. Note that, in this particular situation, the incremental algorithm does not give results that deviate much from the values obtained with the batch learning, providing in some cases even better results. It should also be noted that the results do not improve according to the adaptability of the paradigm to the specific case, since they are better even when the realized one-layer LVM network is probably not suitable for capturing the underlying implication scheme (as seen in the case of Contraceptive Method database).

Finally, the results prove even more interesting if we consider that in both cases (batch and incremental) they are obtained using the same number of epochs (in particular equal to $20$). In fact, an incremental algorithm should typically employ many more steps to achieve the performance of a batch algorithm, whereas in the cases under examination the change seems to bring only advantages.

\section{Discussion and Conclusions}
\label{sec:conc}
In this work, an in-depth analysis of the individual elements necessary to create a Bayesian network through the FGrn paradigm was conducted, showing how it is possible to reduce memory and computational costs during implementation. The analysis led to the creation of a C++ library able to provide excellent results from a computational point of view, transforming polynomial costs into linear (respectively to the number of variables involved). The incremental use of the ML algorithm has finally demonstrated how it is further possible to reduce both the computational and memory costs of the learning phase, even improving, in some of the cases, the ability of the network to learn from the evidence. All these algorithmic choices are at the basis for extending the FGrn paradigm's  to higher-scale problems.

\bibliography{Paper4arXiv}

\end{document}